%% file: main.tex
\documentclass[10pt,twocolumn,letterpaper]{article}

\usepackage[pagenumbers]{wacv} 

\input{preamble}

\definecolor{wacvblue}{rgb}{0.21,0.49,0.74}
\usepackage[pagebackref,breaklinks,colorlinks,allcolors=wacvblue]{hyperref}

\def\wacvPaperID{9} 
\def\confName{HARVEST-Vision}
\def\confYear{2026}

\title{WeedRepFormer: Reparameterizable Vision Transformers for Real-Time Waterhemp Segmentation and Gender Classification}

\author{
Toqi Tahamid Sarker,
Taminul Islam,
Khaled R. Ahmed,\\
Cristiana Bernardi Rankrape,
Kaitlin E. Creager,
Karla Gage\\
Southern Illinois University Carbondale, USA\\
{\tt\small \{toqitahamid.sarker, taminul.islam, khaled.ahmed,}\\
{\tt\small cris.rankrape, kaitlin.creager, kgage\}@siu.edu}
}

\begin{document}
\maketitle
\input{sec/00_abstract}

\input{sec/01_introduction}

\input{sec/02_related_work}

\input{sec/03_architecture}

\input{sec/04_dataset}

\input{sec/05_experiments}
\input{sec/06_conclusion}
{
    \small
    \bibliographystyle{ieeenat_fullname}
    \bibliography{main}
}

\end{document}

%% file: preamble.tex
\usepackage{multirow}
\usepackage{booktabs}
\usepackage{amssymb}
\usepackage{makecell}
\usepackage{float}
\usepackage{makecell} 
\usepackage{colortbl}
\usepackage[table]{xcolor}

%% file: sec/00_abstract.tex
\begin{abstract} 
We present WeedRepFormer, a lightweight multi-task Vision Transformer designed for simultaneous waterhemp segmentation and gender classification. Existing agricultural models often struggle to balance the fine-grained feature extraction required for biological attribute classification with the efficiency needed for real-time deployment. To address this, WeedRepFormer systematically integrates structural reparameterization across the entire architecture—comprising a Vision Transformer backbone, a Lite R-ASPP decoder, and a novel reparameterizable classification head—to decouple training-time capacity from inference-time latency. We also introduce a comprehensive waterhemp dataset containing 10,264 annotated frames from 23 plants. On this benchmark, WeedRepFormer achieves 92.18\% mIoU for segmentation and 81.91\% accuracy for gender classification using only 3.59M parameters and 3.80 GFLOPs. At 108.95 FPS, our model outperforms the state-of-the-art iFormer-T by 4.40\% in classification accuracy while maintaining competitive segmentation performance and significantly reducing parameter count by 1.9$\times$. \end{abstract}

%% file: sec/01_introduction.tex
\section{Introduction}

Waterhemp (\emph{Amaranthus tuberculatus} (Moq.) Sauer) has emerged as one of the most troublesome weed species in North American agriculture, particularly threatening corn and soybean production across the Midwestern United States~\cite{steckel2007dioecious, waselkov2020agricultural, bell2013multiple}. This dioecious species exhibits remarkable adaptability, with separate male and female plants requiring cross-pollination, generating extensive genetic diversity that facilitates rapid evolution of herbicide resistance~\cite{salas2016resistance, ma2015measuring, gage2019emerging}.
Currently, waterhemp populations have developed confirmed resistance to seven herbicide sites of action (SOAs), with an eighth resistance to glufosinate (SOA Group 10) currently being confirmed~\cite{heap2025,rankrape2024evaluating}. This eighth resistance is particularly concerning as glufosinate represents the last remaining postemergence herbicide option in soybean production~\cite{rankrape2024evaluating}. 
The severity of this problem is compounded by the species' exceptional fecundity, with female plants producing between 300,000 to over 2 million seeds per plant~\cite{hartzler2004effect}, contributing to rapid population expansion and yield losses up to 74\% in corn~\cite{steckel2004common} and 43-73\% in soybean~\cite{hager2002common,vyn2007control} under heavy infestation.

The dioecious nature of waterhemp presents both a challenge and an opportunity for precision weed management. Unlike monoecious weeds, waterhemp requires both male and female plants for reproduction, with only female plants producing seeds~\cite{montgomery2019sex}. Male plants produce wind-borne pollen, leading to extensive gene flow that carries herbicide resistance genes up to 800 m from source plants~\cite{liu2012pollen,trucco2011amaranthus}. Recent studies have shown that herbicide exposure can alter population sex ratios~\cite{rumpa2019effect,rumpa2025does}, with PPO-inhibitor treatments shifting male-to-female proportions in ways that could affect seedbank composition and resistance evolution. These dynamics suggest that targeted removal of female plants before seed set could significantly reduce population growth and slow herbicide resistance evolution. However, manual identification of waterhemp gender in field conditions is labor-intensive and impractical at scale, while visual differentiation between male and female plants requires expertise, as morphological differences are subtle and vary with growth stage. Automated gender classification through computer vision could enable selective herbicide application or mechanical removal strategies, potentially reducing chemical inputs while improving long-term management efficacy.

\begin{figure*}[t]
    \centering
    \includegraphics[width=1\linewidth]{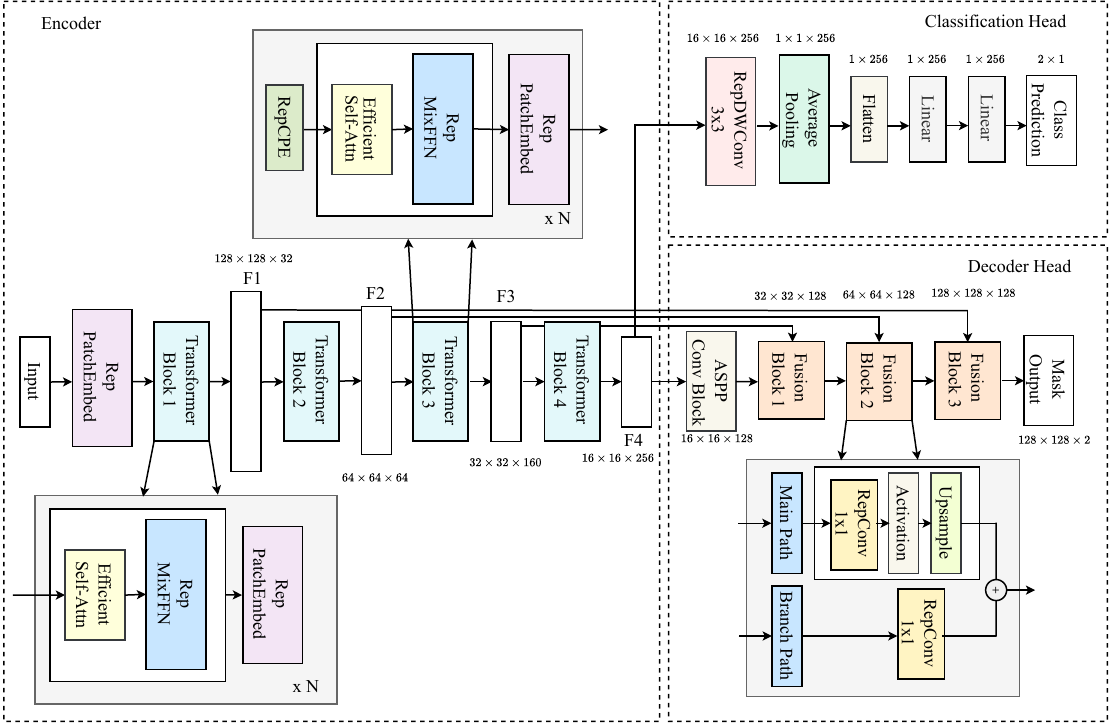}
    \caption{Overview of our multi-task architecture. (a) Network consists of four-stage hierarchical Vision Transformer backbone with reparameterizable components. (b) Reparameterizable patch embedding with multi-branch convolutions. (c) LRASPP decoder with reparameterizable convolutions. (d) Classification head with optional SE attention.}
    \label{fig:arch}
    \vspace{-15pt}
\end{figure*}

Recent advances in deep learning have demonstrated remarkable success in agricultural computer vision tasks~\cite{kamilaris2018deep,saleem2019plant}, with semantic segmentation methods achieving state-of-the-art performance in crop-weed discrimination and identifying weed plant species~\cite{sodjinou2022deep,sarker2025weedsense,islam2025weedswin}. Multi-task learning frameworks that jointly optimize related tasks through shared representations have shown improved efficiency and generalization compared to single-task approaches~\cite{crawshaw2020multi,ruder2017overview}. For waterhemp management, simultaneous segmentation and gender classification are naturally related tasks that benefit from shared feature learning, as both require understanding of plant morphology and spatial structure.

However, practical deployment of deep learning models for agricultural robotics faces significant constraints. Real-time operation requires efficient processing on resource-constrained edge devices such as those mounted on autonomous tractors or unmanned aerial vehicles~\cite{pintus2025emerging}. While Vision Transformers have demonstrated superior performance in semantic segmentation due to their global receptive fields~\cite{xie2021segformer}, their computational cost has limited adoption in resource-constrained agricultural scenarios. Standard models like DeepLabV3+~\cite{chen2018encoder} and U-Net~\cite{ronneberger2015u} achieve high accuracy but require substantial computational resources~\cite{milioto2018real}. Recent lightweight architectures using depthwise separable convolutions~\cite{howard2017mobilenets}, neural architecture search~\cite{tan2019efficientnet}, or knowledge distillation~\cite{hinton2015distilling} often sacrifice accuracy for speed or require complex training procedures.

Structural reparameterization offers a promising alternative, enabling networks to train with multiple parallel branches for increased capacity while deploying as efficient single-path models through algebraic fusion~\cite{ding2021repvgg,vasu2023mobileone}. This approach has achieved state-of-the-art efficiency in image classification and detection~\cite{vasu2023fastvit}, and has been successfully applied to agricultural CNN-based architectures for disease detection, weed identification, and multi-task learning~\cite{zheng2023repdi,pei2025research,guo2025rlcfe,shen2024yolov5}. However, to the best of our knowledge, structural reparameterization has not yet been applied to Vision Transformers for multi-task dense prediction combining segmentation with fine-grained biological attribute classification in agriculture.

We propose WeedRepFormer, a Vision Transformer architecture that achieves an optimal balance of accuracy and efficiency for simultaneous waterhemp segmentation and gender classification via systematic structural reparameterization. In summary, our contributions are as follows:
\begin{itemize}
\setlength\itemsep{-.1em}
    \item We propose a fully reparameterizable multi-task Vision Transformer that systematically applies structural reparameterization across backbone, segmentation head, and classification head.
    \item We introduce a waterhemp dataset with 10,264 annotated frames from 23 plants, including pixel-level segmentation masks and plant-level gender labels, and establish baseline results for this task.

\end{itemize}

%% file: sec/02_related_work.tex
\section{Related Work}

\noindent\textbf{Weed Detection and Segmentation.}
Deep learning has revolutionized weed detection in precision agriculture. Early work adapted general semantic segmentation architectures like FCN~\cite{long2015fully} and U-Net~\cite{ronneberger2015u} for agricultural applications, with successful deployment for real-time crop-weed classification~\cite{lottes2018fully,asad2020weed}. Recent work has explored lightweight architectures for edge devices~\cite{wang2025lightweight,kong2024lightweight}, but these methods primarily address binary crop-weed segmentation without species-specific biological attribute classification. 

\noindent\textbf{Vision Transformers for Segmentation.} Vision Transformers~\cite{dosovitskiy2020image} capture long-range dependencies through self-attention but incur high computational costs. Hierarchical variants address this limitation: SegFormer~\cite{xie2021segformer} combines efficient Transformer encoders with lightweight MLP decoders, while Swin Transformer~\cite{liu2021swin} uses shifted windows for complexity reduction. These architectures have shown success in general computer vision and are increasingly adopted for crop disease detection and weed species classification, though their application to fine-grained multi-task agricultural problems remains limited.

\noindent\textbf{Multi-Task Learning.}
Multi-task learning (MTL) improves efficiency by sharing representations across related tasks~\cite{caruana1997multitask,ruder2017overview}. In agriculture, MTLSegFormer~\cite{goncalves2023mtlsegformer} performs multi-task semantic segmentation with attention-based feature sharing between tasks, while WeedSense~\cite{sarker2025weedsense} jointly performs semantic segmentation, height estimation, and growth stage classification. Other work explores joint classification and regression for hyperspectral images~\cite{chhapariya2025multitask}. These methods demonstrate the benefits of shared feature learning for related agricultural tasks.

\noindent\textbf{Structural Reparameterization.}
Structural reparameterization enables multi-branch training with single-path inference through algebraic fusion. RepVGG~\cite{ding2021repvgg} pioneered this approach for CNNs, MobileOne~\cite{vasu2023mobileone} achieved millisecond-scale mobile latency, and FastViT~\cite{vasu2023fastvit} extended the technique to Vision Transformers achieving significant speedups over hybrid architectures. In agriculture, reparameterization has been successfully applied to CNN-based architectures for disease detection~\cite{zheng2023repdi,pei2025research}, weed detection~\cite{guo2025rlcfe}, instance segmentation~\cite{luo2025enhanced}, and multi-task detection~\cite{shen2024yolov5}. General vision work has also explored reparameterization for multi-task learning~\cite{kanakis2020reparameterizing}. However, existing agricultural applications primarily use CNN-based backbones such as YOLO and RepVGG variants. We extend this paradigm by introducing reparameterizable Vision Transformers for joint segmentation and gender classification, combining the global receptive fields of transformers with the efficiency benefits of structural reparameterization.

%% file: sec/03_architecture.tex
\section{Architecture}
\label{sec:arch}

We propose a fully reparameterizable architecture for real-time waterhemp segmentation and gender classification as shown in ~\Cref{fig:arch}. Following structural reparameterization~\cite{ding2021repvgg,vasu2023mobileone}, we decouple training and inference by using multi-branch overparameterization during training that fuses into efficient single-path inference with zero overhead. Our architecture comprises: (1) a Vision Transformer backbone with RepPatchEmbed, RepCPE, and RepMixFFN, (2) a reparameterizable Lite R-ASPP decoder, and (3) a reparameterizable classification head.

\begin{figure}[tbp]
    \centering
    \begin{subfigure}[b]{0.44\textwidth}
        \centering
        \includegraphics[width=\linewidth]{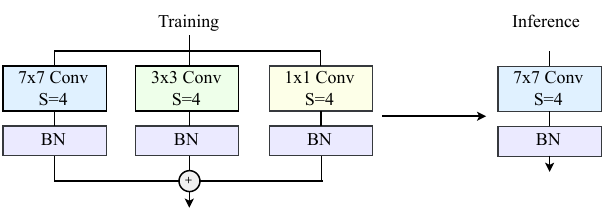}
        \caption{RepPatchEmbed}
        \label{fig:reppatchembed}
    \end{subfigure}
    \\ \vspace{2pt}
    \begin{subfigure}[b]{0.44\textwidth}
        \centering
        \includegraphics[width=\linewidth]{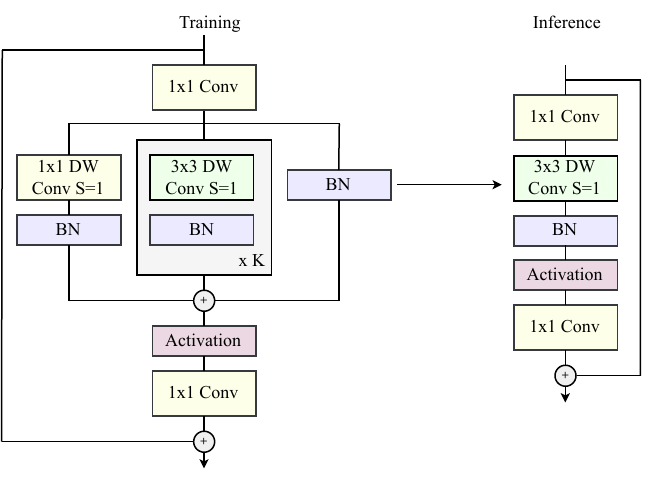}
        \caption{RepMixFFN}
        \label{fig:repmixffn}
    \end{subfigure}
    \\ \vspace{2pt}
    \begin{subfigure}[b]{0.24\textwidth}
        \centering
        \includegraphics[width=\linewidth]{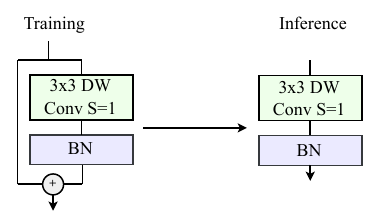}
        \caption{RepCPE}
        \label{fig:repcpe}
    \end{subfigure}
    \vspace{-5pt}
     \caption{Reparameterizable components with train-time overparameterization. \subref{fig:reppatchembed} RepPatchEmbed uses three parallel branches that fuse at inference. \subref{fig:repmixffn} RepMixFFN employs K parallel depthwise convolutions with identity connection. \subref{fig:repcpe} RepCPE combines depthwise convolution with identity for positional encoding.}
    \label{fig:blocks}
    \vspace{-15pt}
\end{figure}

\subsection{Vision Transformer Backbone}

We extend the hierarchical Mix Transformer (MiT) backbone from SegFormer~\cite{xie2021segformer} with systematic reparameterization across all components. The backbone consists of four stages with progressively downsampled feature maps at resolutions $\{\frac{H}{4}, \frac{H}{8}, \frac{H}{16}, \frac{H}{32}\}$, producing multi-scale features $\{\mathbf{F}_1, \mathbf{F}_2, \mathbf{F}_3, \mathbf{F}_4\}$ with channel dimensions $[32, 64, 160, 256]$. We use efficient multi-head self-attention with spatial reduction ratios $r \in \{8,4,2,1\}$ across stages to reduce complexity at higher resolutions.

\noindent\textbf{RepPatchEmbed.} Following MobileOne~\cite{vasu2023mobileone}, we use train-time overparameterization for patch embedding layers as shown in ~\Cref{fig:reppatchembed}. RepPatchEmbed employs three parallel branches: a large kernel conv ($p_i \times p_i$ where $p_i \in \{7,3,3,3\}$ for stages 1-4), a $3 \times 3$ conv, and a $1 \times 1$ conv. The multi-scale branches improve low-level feature learning. At deployment, all branches fuse into a single $p_i \times p_i$ convolution with zero computational overhead.

\noindent\textbf{RepCPE.} Unlike absolute positional encodings, conditional positional encodings (CPE)~\cite{chu2021conditional,chu2021twins} are dynamically generated based on local input context. We apply reparameterizable CPE~\cite{vasu2023fastvit} in stages 3--4 as shown in ~\Cref{fig:repcpe}. RepCPE adds a $3 \times 3$ depthwise convolution to the input: $\text{RepCPE}(\mathbf{F}) = \mathbf{F} + \text{DWConv}_{3\times3}(\mathbf{F})$. At inference, the identity and convolution fuse into a single operation.

\noindent\textbf{RepMixFFN.} We enhance the MixFFN with multi-branch depthwise convolutions for spatial mixing as shown in ~\Cref{fig:repmixffn}. RepMixFFN uses $K$ parallel $3 \times 3$ depthwise branches plus a $1 \times 1$ branch and identity connection between two pointwise layers. Layer scale~\cite{liu2022convnext} initialized to $10^{-5}$ stabilizes training. All branches fuse at deployment for efficient inference.

\begin{figure}[tbp]
    \centering
    \begin{subfigure}[b]{0.40\textwidth}
        \centering
        \includegraphics[width=\linewidth]{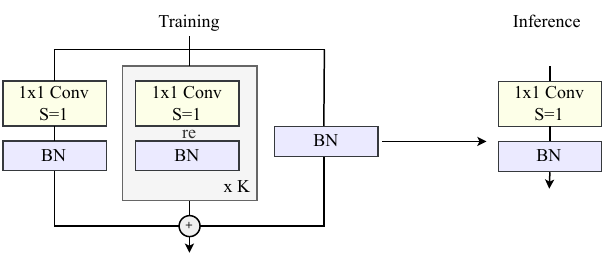}
        \caption{RepConv1x1}
        \label{fig:repconv1x1}
    \end{subfigure}
    \\ \vspace{5pt}
    \begin{subfigure}[b]{0.40\textwidth}
        \centering
        \includegraphics[width=\linewidth]{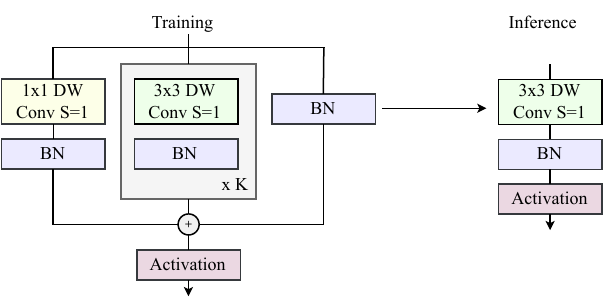}
        \caption{RepDWConv3x3}
        \label{fig:repdwconv3x3}
    \end{subfigure}
    \vspace{-5pt}
    \caption{Reparameterizable convolution modules used in decoder and classifier. \subref{fig:repconv1x1} RepConv1x1 with K parallel 1×1 branches for efficient channel mixing. \subref{fig:repdwconv3x3} RepDWConv3x3 with K parallel 3×3 depthwise branches for spatial feature refinement.}
    \label{fig:modules}
    \vspace{-10pt}
\end{figure}

\subsection{Segmentation Decoder and Classification Head}

\noindent\textbf{RepLR-ASPP Decoder.} For efficient multi-scale feature aggregation, we adapt Lite R-ASPP~\cite{howard2019searching} by replacing all convolutions with reparameterizable blocks. LR-ASPP provides a lightweight alternative to ASPP~\cite{chen2017deeplab} by using global pooling instead of dilated convolutions. The decoder uses RepConv1x1 blocks as shown in ~\Cref{fig:repconv1x1}, which employ $K$ parallel 1×1 branches with layer scale that fuse at deployment. We first apply global context attention to $\mathbf{F}_4$ via global average pooling and sigmoid gating. Then, features are progressively upsampled and fused with lateral connections from $\{\mathbf{F}_1, \mathbf{F}_2, \mathbf{F}_3\}$ through channel concatenation and RepConv1x1 fusion to produce the final segmentation output.

\noindent\textbf{RepClsHead.} We propose RepClsHead, a reparameterizable classification head inspired by FastViT~\cite{vasu2023fastvit}. Following FastViT's design, RepClsHead refines features before global pooling using RepDWConv3x3 as shown in ~\Cref{fig:repdwconv3x3}, which employs $K$ parallel $3 \times 3$ depthwise branches plus a $1 \times 1$ depthwise branch and identity connection. After refinement, features undergo global average pooling followed by a two-layer MLP with hidden dimension 256 and dropout rate 0.5 to produce binary gender classification logits. All branches fuse into a single $3 \times 3$ depthwise convolution at deployment.

\noindent\textbf{Multi-Task Loss.} \label{sec:multitask_loss} We jointly optimize the network for segmentation and classification using a multi-task loss $\mathcal{L} = \mathcal{L}_{\text{seg}} + \lambda \mathcal{L}_{\text{cls}}$, where $\lambda = 0.5$. $\mathcal{L}_{\text{seg}}$ denotes pixel-wise cross-entropy over background and plant classes, while $\mathcal{L}_{\text{cls}}$ is binary cross-entropy for gender classification.

%% file: sec/04_dataset.tex
\section{Dataset}
\label{sec:dataset}

\begin{figure}[t] 
\centering
    
    \includegraphics[width=0.108\textwidth]{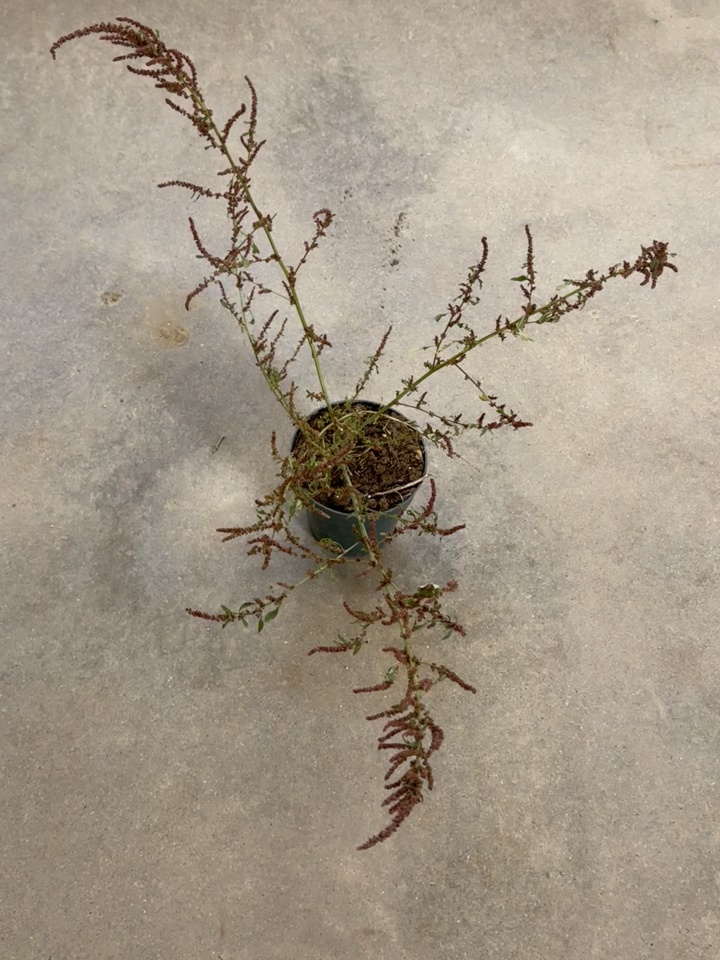}
    \includegraphics[width=0.108\textwidth]{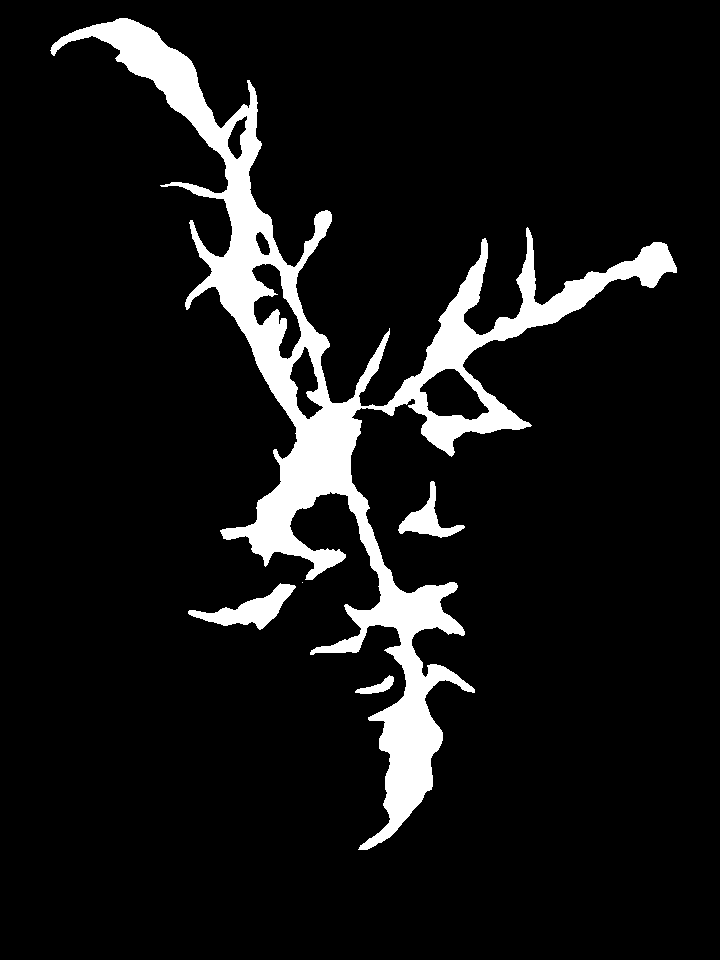}
    \includegraphics[width=0.108\textwidth]{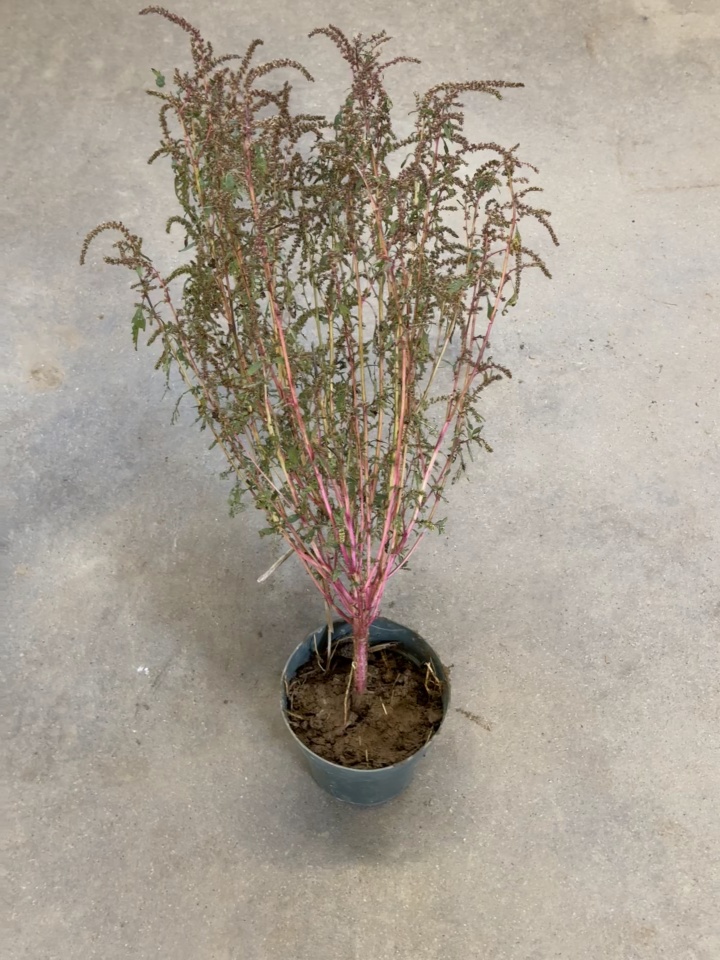}
    \includegraphics[width=0.108\textwidth]{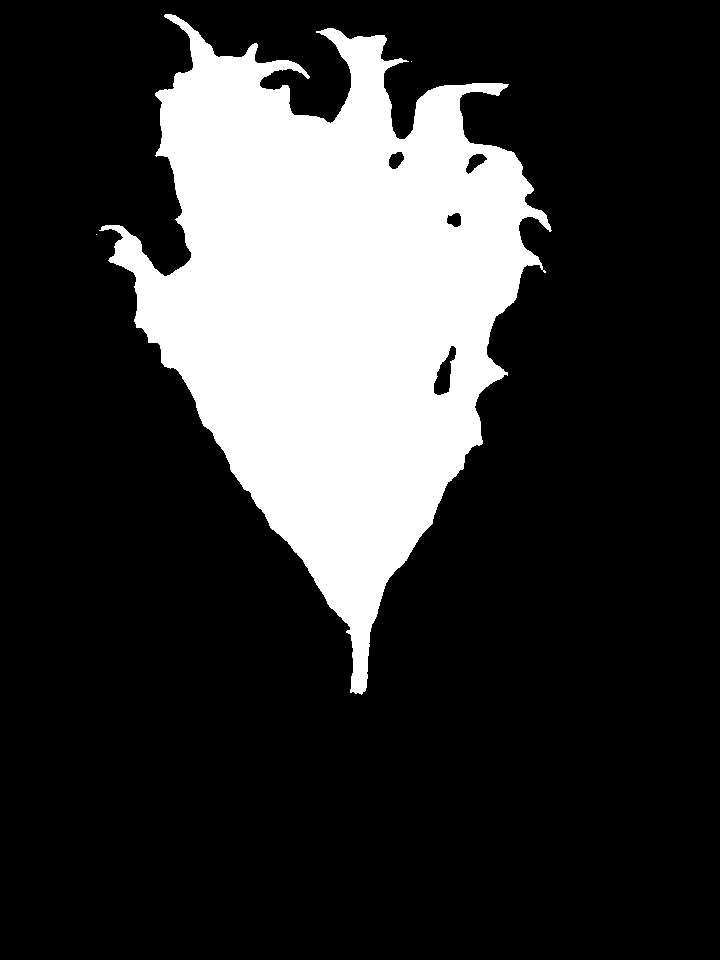}
    \\
    \makebox[0.108\textwidth]{\scriptsize Image}
    \makebox[0.108\textwidth]{\scriptsize Mask}
    \makebox[0.108\textwidth]{\scriptsize Image}
    \makebox[0.108\textwidth]{\scriptsize Mask}
    \\
    \caption{Example image-mask pairs from the waterhemp dataset. Left pair shows a female specimen, right pair shows a male specimen. Binary segmentation masks delineate complete plant architecture against background.}
  \label{fig:dataset_samples}
  \vspace{-8pt}
\end{figure}

We introduce a novel waterhemp plant dataset designed for joint semantic segmentation and gender classification tasks. 
The dataset comprises 23 individual mature waterhemp plants, specifically 13 females and 10 males identified with the assistance of an agricultural expert, which were collected from a field site and transplanted into pots for imaging.
All plants were at the flowering stage, exhibiting fully developed morphological characteristics essential for gender differentiation. The complete dataset statistics are presented in Table~\ref{tab:dataset_stats}. Representative samples from the dataset are shown in Figure~\ref{fig:dataset_samples}.

\subsection{Plant Collection}

Waterhemp plants were collected from Field 27 at SIUC University Farms (37.706°N, 89.253°W) on October 14, 2024 and transplanted into 6" diameter plastic pots for imaging.
The field had been fertilized with 200 lbs per acre of potash in spring 2024. Soybeans (Pioneer P37A18E) were planted on June 13, 2024 at 180,000 seeds per acre with 15-inch row spacing. The field was tilled at planting to remove weed competition. A single early postemergence herbicide application was made on August 12, 2024, consisting of glufosinate (Liberty, 32 oz/acre), glyphosate (Roundup Powermax 3, 1 qt/acre), citric acid (2.4 oz/acre), and ammonium sulfate (2.4 lb/acre). Soybeans were harvested on October 21, 2024 with a yield of approximately 25 bu/acre.

\subsection{Data Collection and Preprocessing}

We captured one video per plant using an iPhone 15 Pro Max, performing a complete 360-degree horizontal rotation to ensure comprehensive morphological coverage. To balance dataset size with temporal diversity, we sample every third frame, effectively reducing redundancy while preserving motion variation. Frames are resized to $720\times960$ pixels, maintaining sufficient spatial resolution for fine-grained structural analysis while halving storage requirements. The final dataset comprises 10,264 frames across 23 plants.

\subsection{Annotation Protocol}

We employ SAM2.1 Hiera Large~\cite{ravi2024sam} for efficient semi-automatic mask generation. For each video sequence, we manually annotate positive and negative prompt points on the initial frame using an interactive bounding box interface. SAM2.1 then propagates these sparse annotations across all subsequent frames via its video predictor, ensuring temporal consistency. The resulting binary masks delineate the complete plant structure (stems, leaves, flowers) as foreground, with all other regions classified as background. Gender labels are assigned at the plant level based on botanical verification during the flowering stage.

\subsection{Dataset Split Strategy}

To prevent data leakage, we partition the dataset at the plant level rather than the image level. Specifically, all frames extracted from a single plant are assigned exclusively to one split—training, validation, or test. This ensures that no visual information from an individual plant appears across multiple splits, thereby providing a more realistic evaluation of model generalization to unseen specimens. Unlike image-level splitting, which can inadvertently leak visual patterns across splits when consecutive frames share similar appearances, our plant-level strategy guarantees complete independence between training and evaluation data.

The plant-level split allocation is as follows: 15 plants for training (9 female, 6 male), 4 plants for validation (2 female, 2 male), and 4 plants for testing (2 female, 2 male). Table~\ref{tab:dataset_stats} summarizes the resulting frame-level distribution across splits. This yields a split ratio of approximately 70.5\% for training, 15.5\% for validation, and 13.9\% for testing. The dataset exhibits a natural gender imbalance, with female plants contributing 59.5\% of total frames. This distribution reflects inherent biological differences: female waterhemp plants typically develop denser foliage and more complex branching structures compared to their male counterparts, resulting in longer video sequences and more extracted frames per plant.

\begin{table}[t]
\centering
\scriptsize
\setlength{\tabcolsep}{5pt}
\begin{tabular}{@{}lcccccc@{}}
\toprule
\textbf{Gender Type} & \textbf{Frames} & \textbf{Percentage} & \textbf{Plants} & \textbf{\makecell{Train\\(70.5\%)}} & \textbf{\makecell{Val\\(15.5\%)}} & \textbf{\makecell{Test\\(13.9\%)}} \\
\midrule
Male & 4,158 & 40.5\% & 10 & 2,692 & 780 & 686 \\
Female & 6,106 & 59.5\% & 13 & 4,544 & 816 & 746 \\
\midrule
\textbf{Total} & \textbf{10,264} & \textbf{100.0\%} & \textbf{23} & \textbf{7,236} & \textbf{1,596} & \textbf{1,432} \\
\bottomrule
\end{tabular}
\caption{Waterhemp dataset statistics. Percentages show distribution within 10,264 annotated frames.}
\label{tab:dataset_stats}
\vspace{-10pt}
\end{table}

%% file: sec/05_experiments.tex
\section{Experiments}
\label{sec:experiments}

\begin{table*}[t]
\centering
\scriptsize
\setlength{\tabcolsep}{8pt}
\begin{tabular}{@{}l|cc|cc|ccc@{}}
\toprule
\multirow{3}{*}{\textbf{Method}} & \multicolumn{2}{c|}{\textbf{Segmentation}} & \multicolumn{2}{c|}{\textbf{Classification}} & \multicolumn{3}{c}{\textbf{Efficiency}} \\
\cmidrule(lr){2-3} \cmidrule(lr){4-5} \cmidrule(lr){6-8}
 & \textbf{mIoU (\%)$\uparrow$} & \textbf{mFscore (\%)$\uparrow$} & \textbf{mAcc (\%)$\uparrow$} & \textbf{mF1 (\%)$\uparrow$} & \textbf{FPS$\uparrow$} & \textbf{GFLOPs$\downarrow$} & \textbf{Params (M)$\downarrow$} \\
\midrule
\multicolumn{8}{l}{\textit{Conv-based Models}} \\
\midrule
BiSeNet~\cite{yu2018bisenet} & 92.27 & 95.87 & 67.25 & 66.87 & \underline{233.45} & 14.821 & 13.455 \\
BiSeNet V2~\cite{yu2021bisenet} & 92.30 & 95.89 & 58.17 & 57.96 & 159.61 & 12.286 & 14.820 \\
DDRNet~\cite{pan2022deep} & 92.81 & 96.18 & 45.74 & 44.94 & 151.58 & 4.560 & 5.766 \\
Fast-SCNN~\cite{poudel2019fast} & 88.27 & 93.51 & 59.57 & 59.41 & 225.70 & \textbf{0.927} & \textbf{1.488} \\
ICNet~\cite{zhao2018icnet} & 91.58 & 95.48 & 56.08 & 56.02 & 134.73 & 15.426 & 47.859 \\
MobileNetV2~\cite{sandler2018mobilenetv2} & 91.50 & 95.43 & 70.88 & 69.34 & 179.47 & 39.261 & 9.793 \\
MobileNetV3~\cite{howard2019searching} & 92.28 & 95.88 & 68.99 & 67.92 & 121.55 & 8.692 & 3.529 \\
MobileOne-S0~\cite{vasu2023mobileone} & 73.02 & 82.76 & 60.61 & 59.23 & \textbf{296.97} & 8.213 & 28.915 \\
RepViT-M0.9~\cite{wang2024repvit} & \textbf{94.31} & \textbf{97.01} & 69.20 & 68.86 & 82.23 & 25.404 & 8.954 \\
\midrule
\multicolumn{8}{l}{\textit{Transformer-based Models}} \\
\midrule
SegFormer-B0~\cite{xie2021segformer} & \underline{94.10} & \underline{96.90} & 74.72 & 75.34 & 115.27 & 7.885 & 3.782 \\
iFormer-T~\cite{zheng2025iformer} & 94.05 & 96.87 & \underline{77.51} & \underline{77.08} & 99.05 & 24.267 & 6.804 \\
FastViT-T8~\cite{vasu2023fastvit} & 93.33 & 96.47 & 55.24 & 54.41 & 189.71 & 23.806 & 7.382 \\
\midrule
\rowcolor{gray!20}
\textbf{WeedRepFormer (Ours)} & 92.18 & 95.82 & \textbf{81.91} & \textbf{81.90} & 108.95 & \underline{3.801} & \underline{3.592} \\
\bottomrule
\end{tabular}
\vspace{-5pt}
\caption{Comparison with state-of-the-art methods on waterhemp multi-task learning. All models trained with identical protocols. Best results in \textbf{bold}, second best \underline{underlined}.}
\label{tab:comparison}
\end{table*}

\noindent\textbf{Implementation Details.} All models are trained for 80K iterations using AdamW optimizer with learning rate $6 \times 10^{-5}$, weight decay $0.01$, and momentum $(0.9, 0.999)$. The learning rate follows a polynomial decay (power 1.0) after a 1,500-iteration linear warmup with start factor $10^{-6}$. We use batch size 8 on an NVIDIA A100 80GB PCIe GPU at $512 \times 512$ resolution with standard augmentations including random horizontal flip, resize, and photometric distortion. The multi-task loss uses $\lambda=0.5$ as described in Section~\ref{sec:multitask_loss}. For fair comparison, all comparison methods utilize a standardized MLP classification head comprising global pooling followed by two linear layers with Xavier initialization, except Fast-SCNN which loads Cityscapes weights~\cite{poudel2019fast}. Our WeedRepFormer employs the proposed RepClsHead and initializes from SegFormer-B0 ImageNet weights~\cite{xie2021segformer}, where standard components transfer directly, while reparameterizable modules inherit weights into their primary branch matching the source kernel size, with additional branches using Kaiming initialization~\cite{he2015delving}. The backbone learning rate is scaled by $0.1\times$ while the classification head uses $1.0\times$ the base rate for rapid adaptation. We report mean IoU (mIoU) and mean F-score (mFscore) for segmentation; mean accuracy (mAcc) and mean F1 (mF1) for classification; and inference FPS.

\subsection{Comparison with State-of-the-Art Methods}

We compare our WeedRepFormer against state-of-the-art efficient segmentation and multi-task models on the waterhemp dataset. Table~\ref{tab:comparison} presents comprehensive results across segmentation, classification, and efficiency metrics.

\noindent\textbf{Classification Performance.} WeedRepFormer achieves 81.91\% mAcc and 81.90\% mF1, outperforming all other methods on gender classification. Compared to the second-best method iFormer-T with 77.51\% mAcc, our approach improves accuracy by 4.40\% and F1 score by 4.82\%. Among transformer baselines, WeedRepFormer surpasses SegFormer-B0 by 7.19\% mAcc and RepViT-M0.9 by 12.71\% mAcc. In contrast, convolution-based models struggle with this task; for instance, BiSeNet and DDRNet achieve only 67.25\% and 45.74\% mAcc respectively, despite their competitive segmentation results. These findings suggest that our structural reparameterization approach effectively captures the discriminative features required for fine-grained classification.

\noindent\textbf{Segmentation Performance.} On segmentation, WeedRepFormer yields 92.18\% mIoU and 95.82\% mFscore. RepViT-M0.9 achieves the highest segmentation at 94.31\% mIoU and 97.01\% mFscore, followed by SegFormer-B0 at 94.10\% mIoU and 96.90\% mFscore. However, these methods require substantially higher computation while achieving weaker classification performance. Among efficient models under 5 GFLOPs, WeedRepFormer achieves competitive segmentation accuracy, comparable to DDRNet (92.81\% mIoU) which drastically fails at classification (45.74\% mAcc). The modest 2.13\% mIoU gap compared to RepViT-M0.9 represents a favorable trade-off for superior classification and efficiency.

\newcommand{\qualfigsize}{0.105}
\newcommand{\quallabelwidth}{0.02}
\newcommand{\qualhspacing}{-1.5pt}

\begin{figure*}[t] 
    \centering
    
    \makebox[\quallabelwidth\textwidth][c]{\scriptsize}
    \makebox[\qualfigsize\textwidth][c]{\scriptsize Input Image}\hspace{\qualhspacing}
    \makebox[\qualfigsize\textwidth][c]{\scriptsize Ground Truth}\hspace{\qualhspacing}
    \makebox[\qualfigsize\textwidth][c]{\scriptsize RepViT}\hspace{\qualhspacing}
    \makebox[\qualfigsize\textwidth][c]{\scriptsize FastViT}\hspace{\qualhspacing}
    \makebox[\qualfigsize\textwidth][c]{\scriptsize MobileOne}\hspace{\qualhspacing}
    \makebox[\qualfigsize\textwidth][c]{\scriptsize MobileNetV2}\hspace{\qualhspacing}
    \makebox[\qualfigsize\textwidth][c]{\scriptsize MobileNetV3}\hspace{\qualhspacing}
    \makebox[\qualfigsize\textwidth][c]{\scriptsize Segformer}\hspace{\qualhspacing}
    \makebox[\qualfigsize\textwidth][c]{\scriptsize WeedRepFormer}
    \\[0pt]

     \raisebox{3ex}{\rotatebox[origin=c]{90}{\textbf{Male}}}\hspace{0pt}
    \includegraphics[width=\qualfigsize\textwidth]{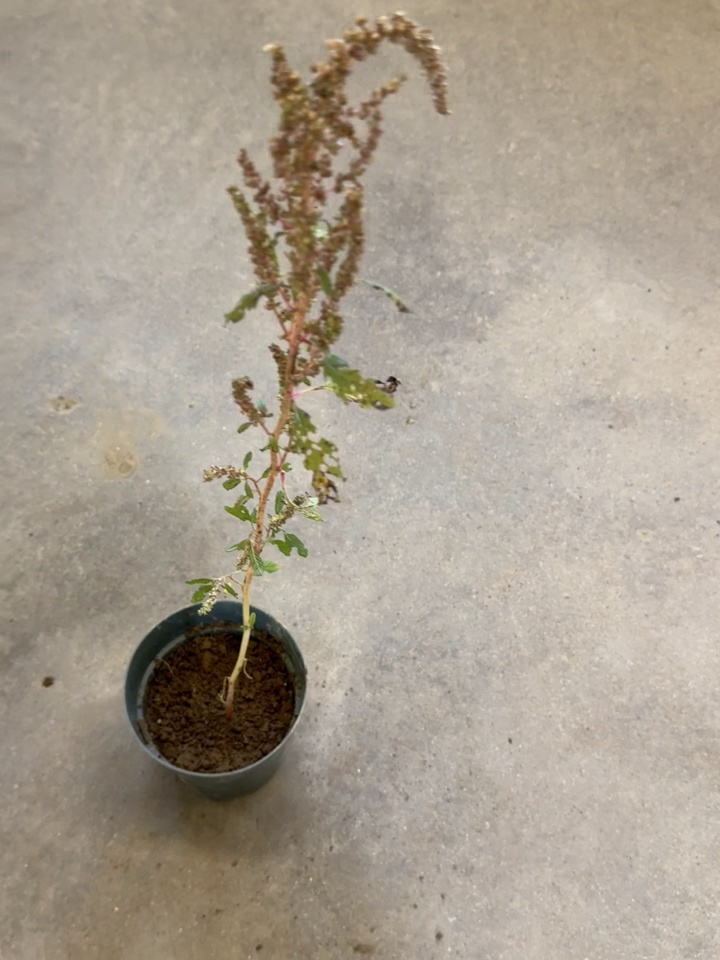}\hspace{\qualhspacing}
    \includegraphics[width=\qualfigsize\textwidth]{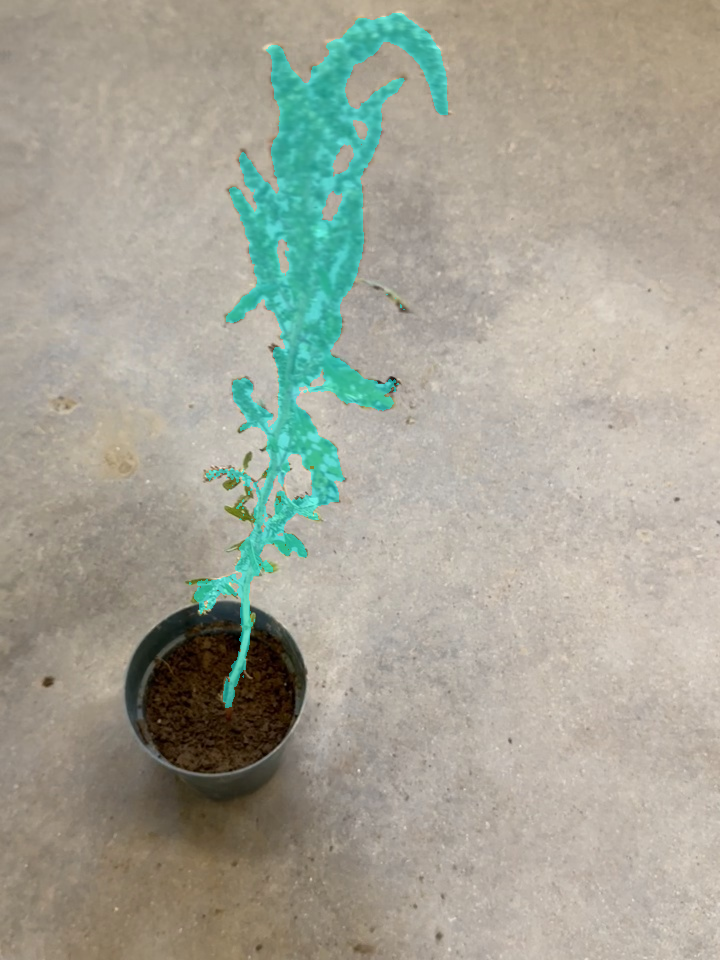}\hspace{\qualhspacing}
    \includegraphics[width=\qualfigsize\textwidth]{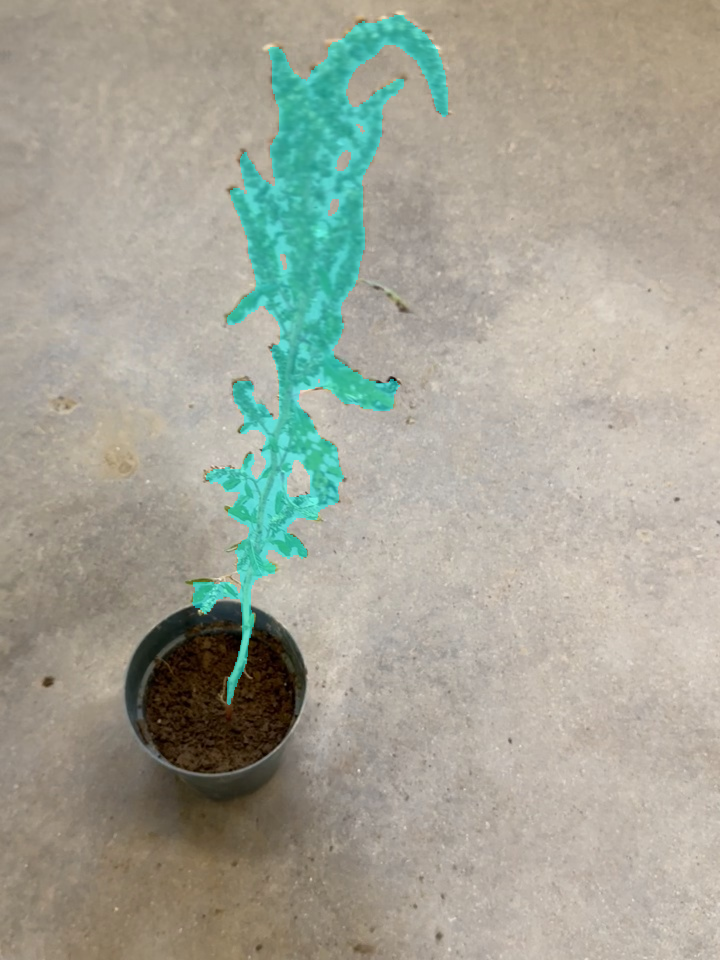}\hspace{\qualhspacing}
    \includegraphics[width=\qualfigsize\textwidth]{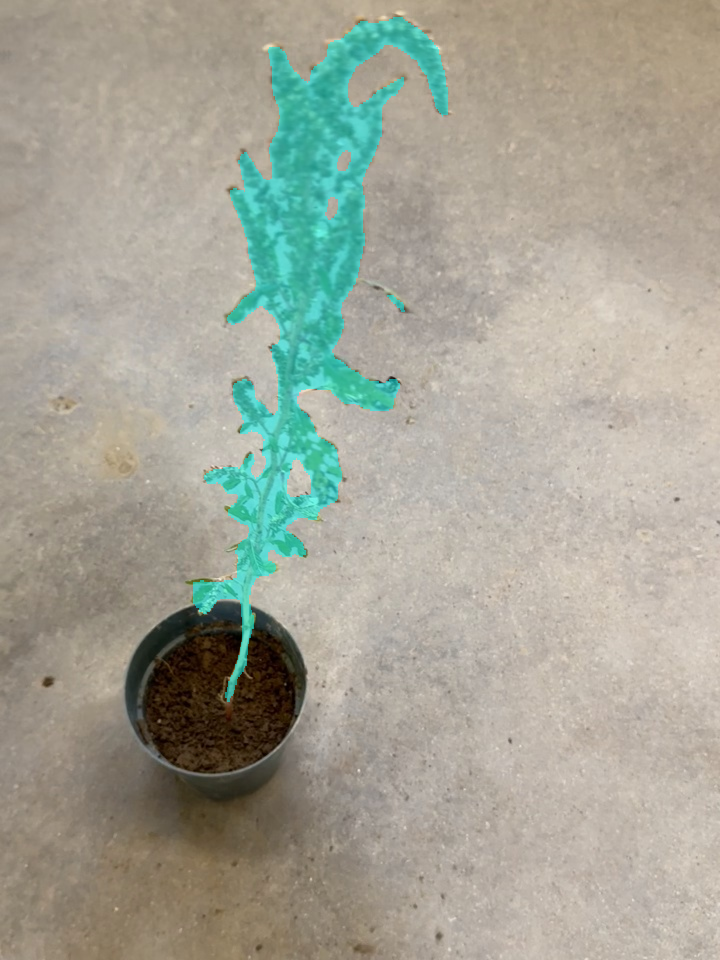}\hspace{\qualhspacing}
    \includegraphics[width=\qualfigsize\textwidth]{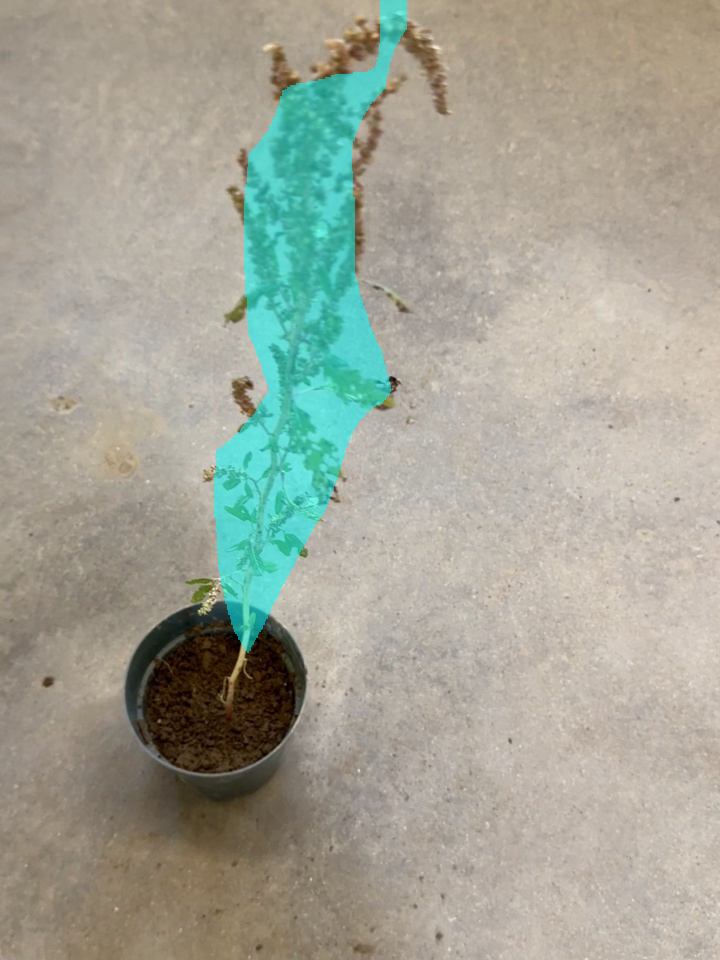}\hspace{\qualhspacing}
    \includegraphics[width=\qualfigsize\textwidth]{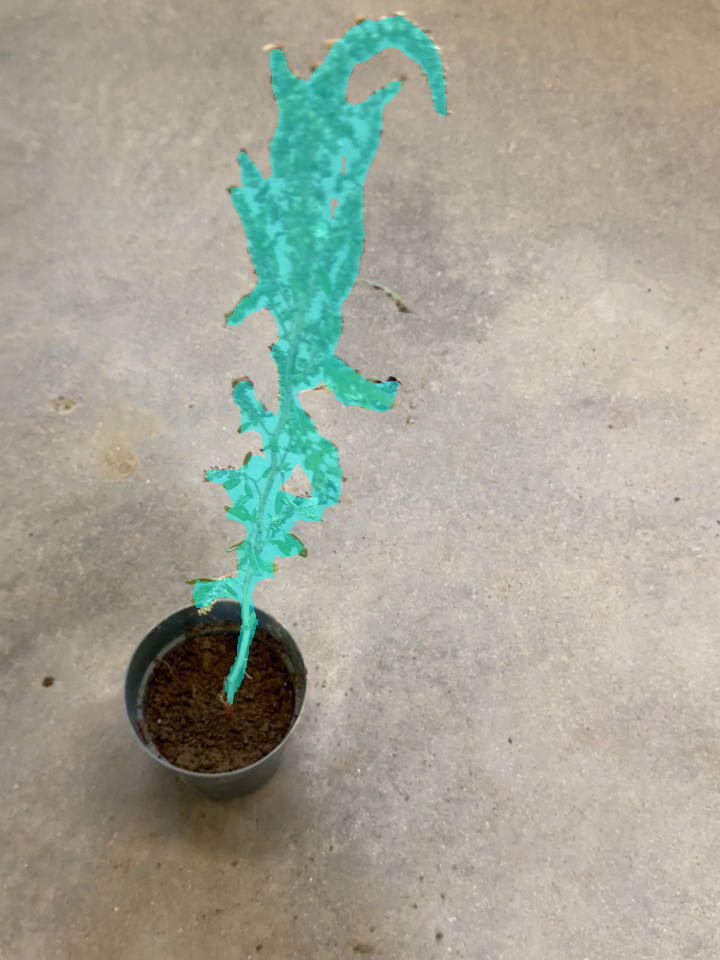}\hspace{\qualhspacing}
    \includegraphics[width=\qualfigsize\textwidth]{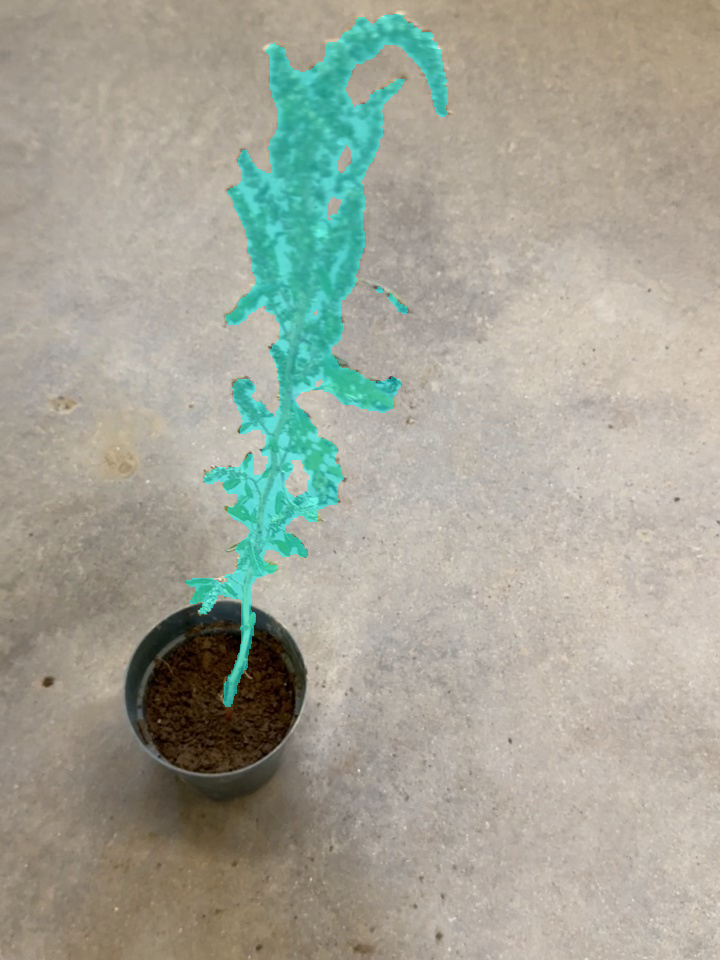}\hspace{\qualhspacing}
    \includegraphics[width=\qualfigsize\textwidth]{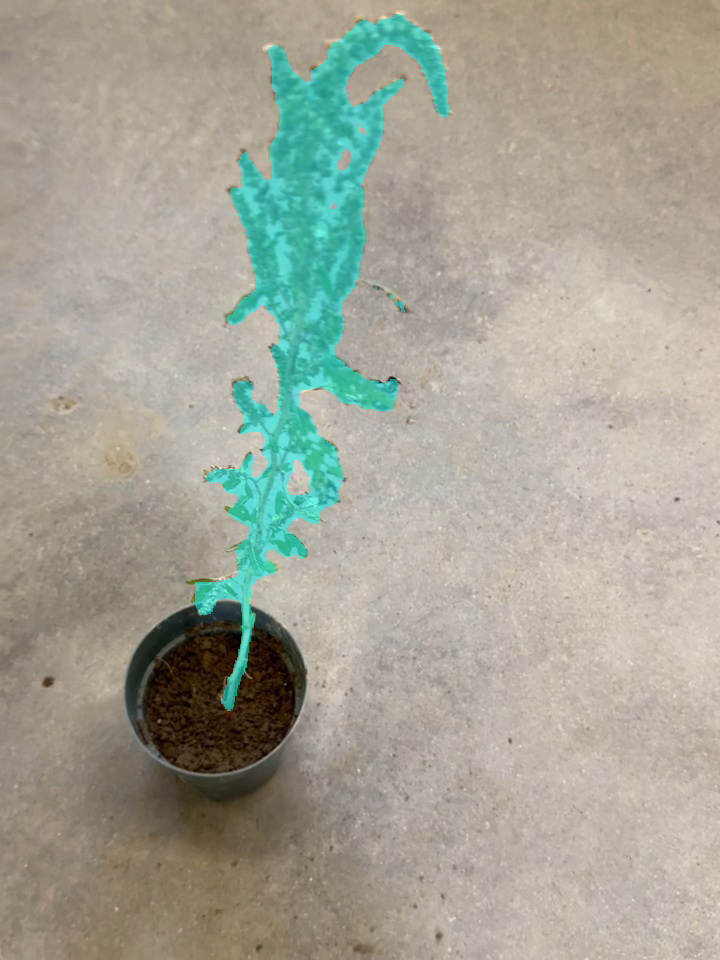}\hspace{\qualhspacing}
    \includegraphics[width=\qualfigsize\textwidth]{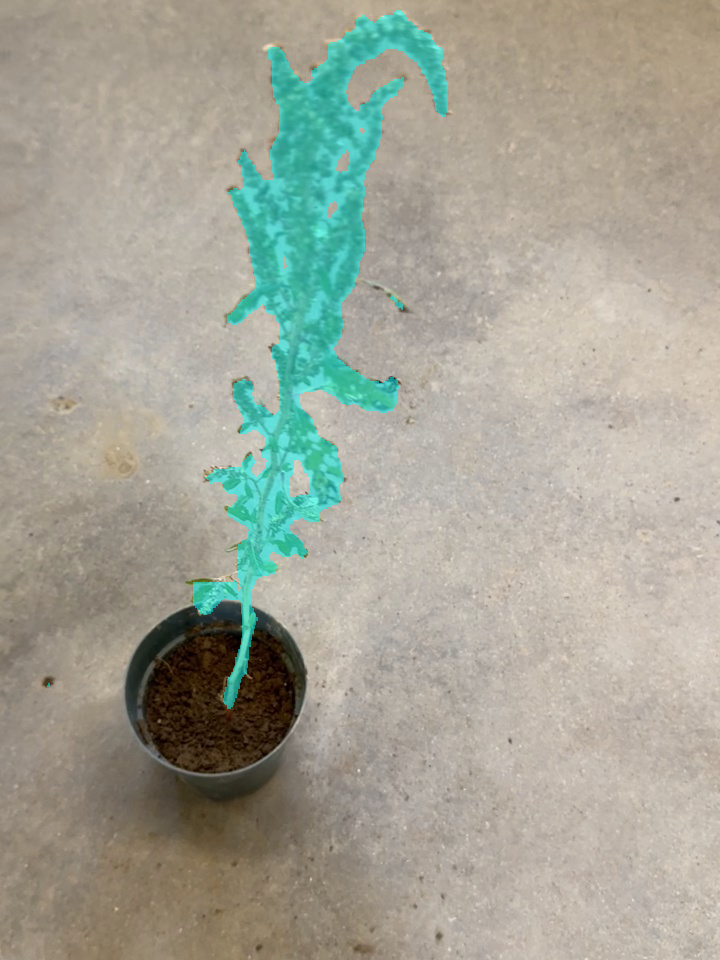}
    \\[2pt]

    \raisebox{3.5ex}{\rotatebox[origin=c]{90}{\textbf{Female}}}\hspace{0pt}
    \includegraphics[width=\qualfigsize\textwidth]{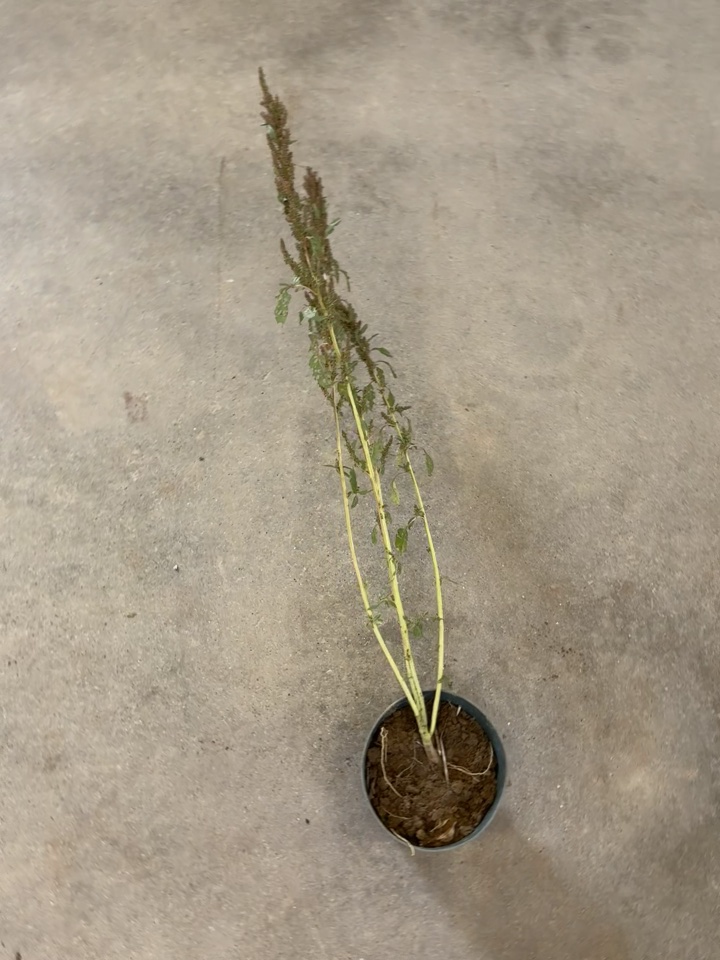}\hspace{\qualhspacing}
    \includegraphics[width=\qualfigsize\textwidth]{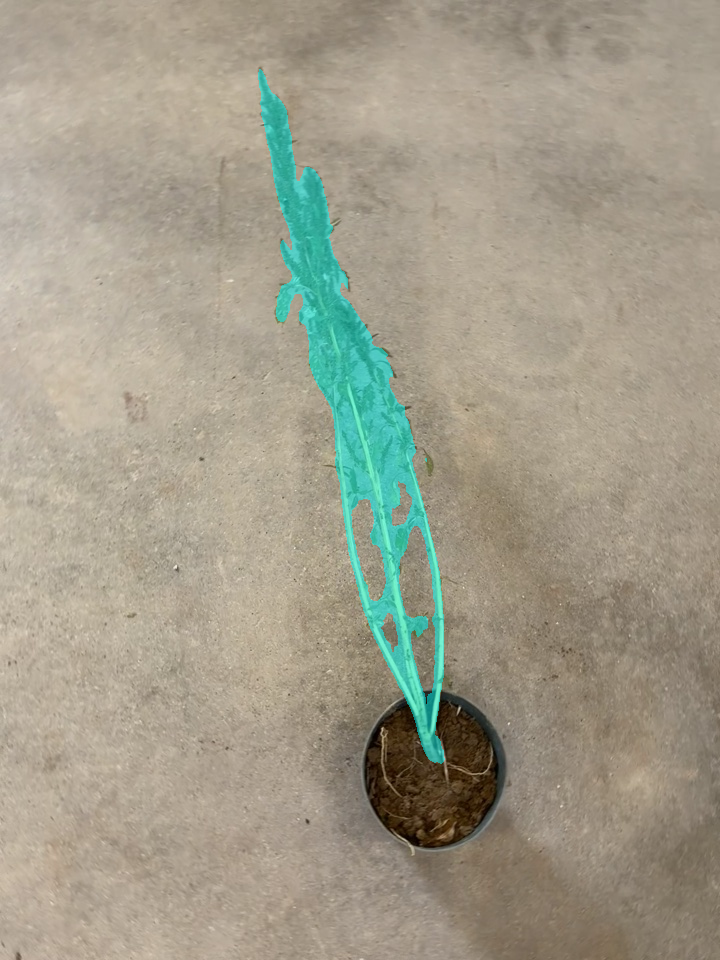}\hspace{\qualhspacing}
    \includegraphics[width=\qualfigsize\textwidth]{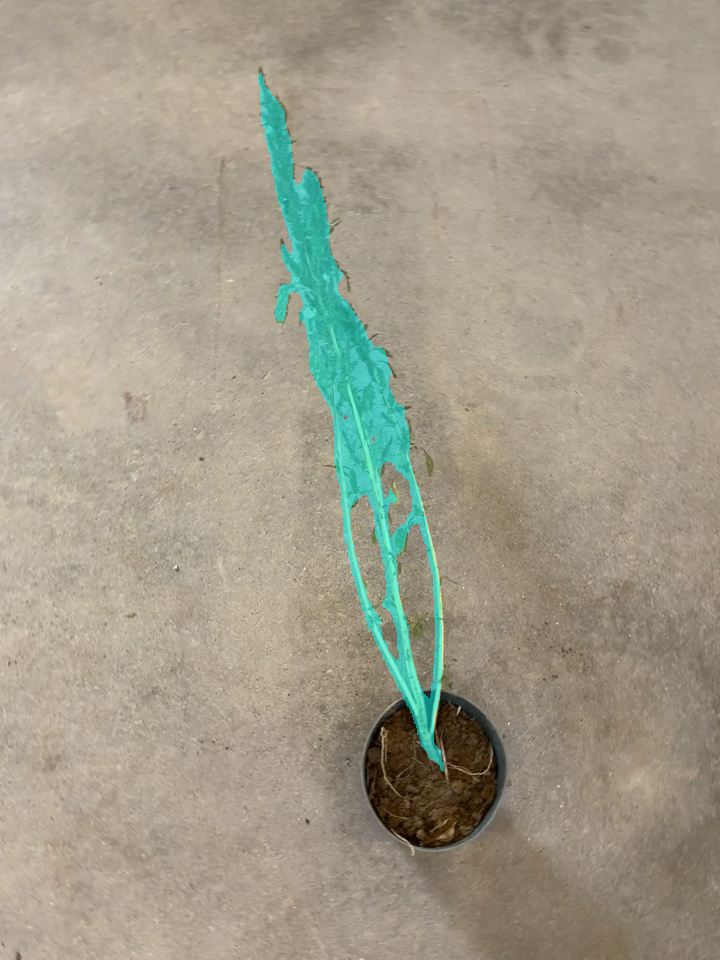}\hspace{\qualhspacing}
    \includegraphics[width=\qualfigsize\textwidth]{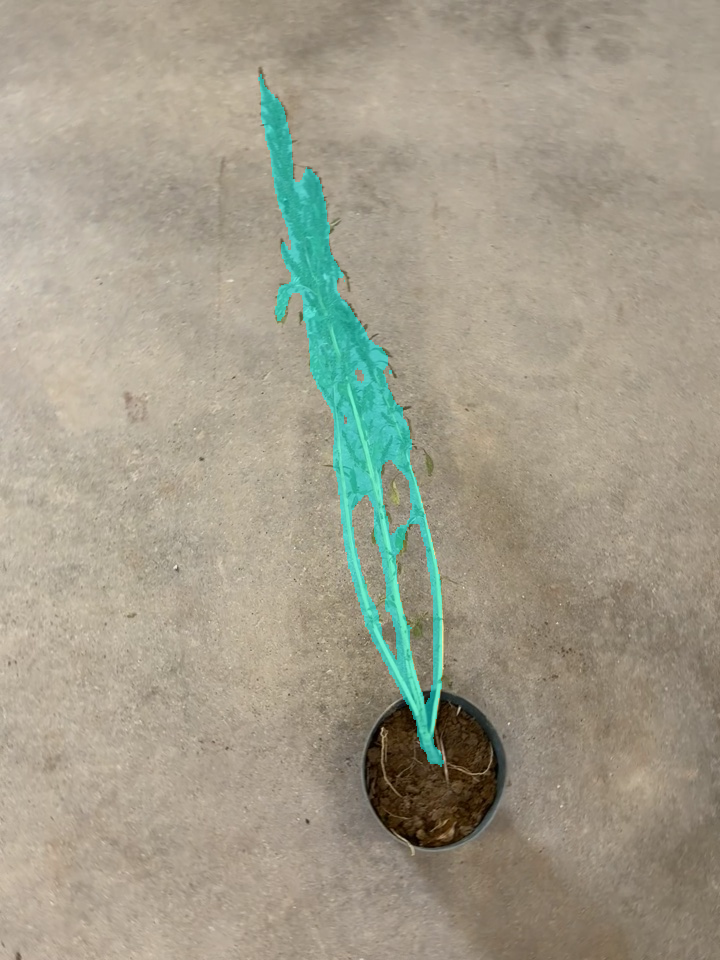}\hspace{\qualhspacing}
    \includegraphics[width=\qualfigsize\textwidth]{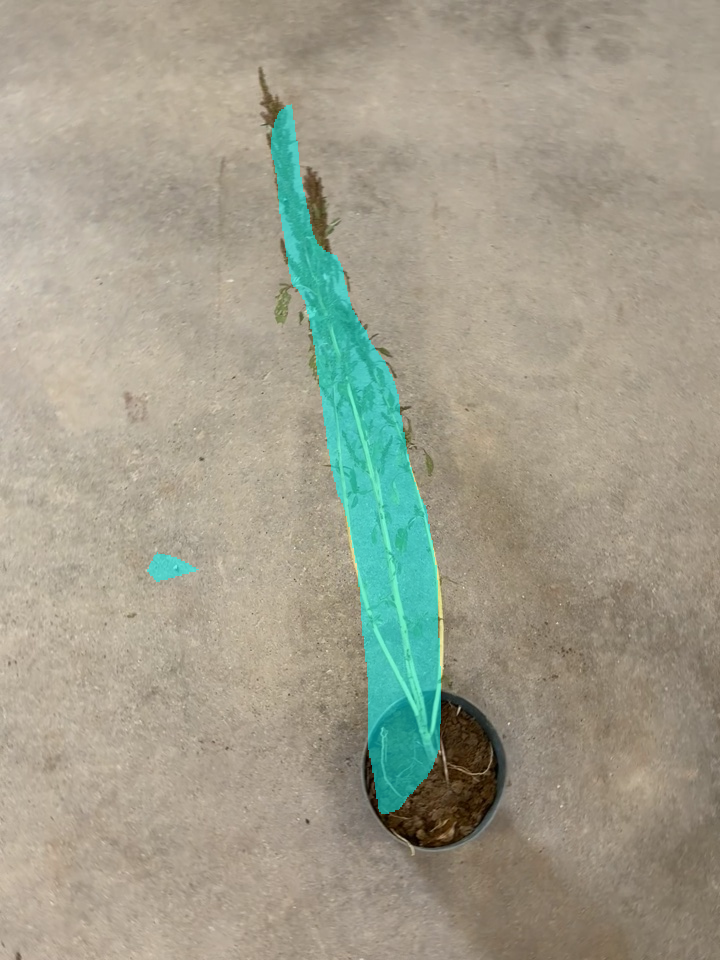}\hspace{\qualhspacing}
    \includegraphics[width=\qualfigsize\textwidth]{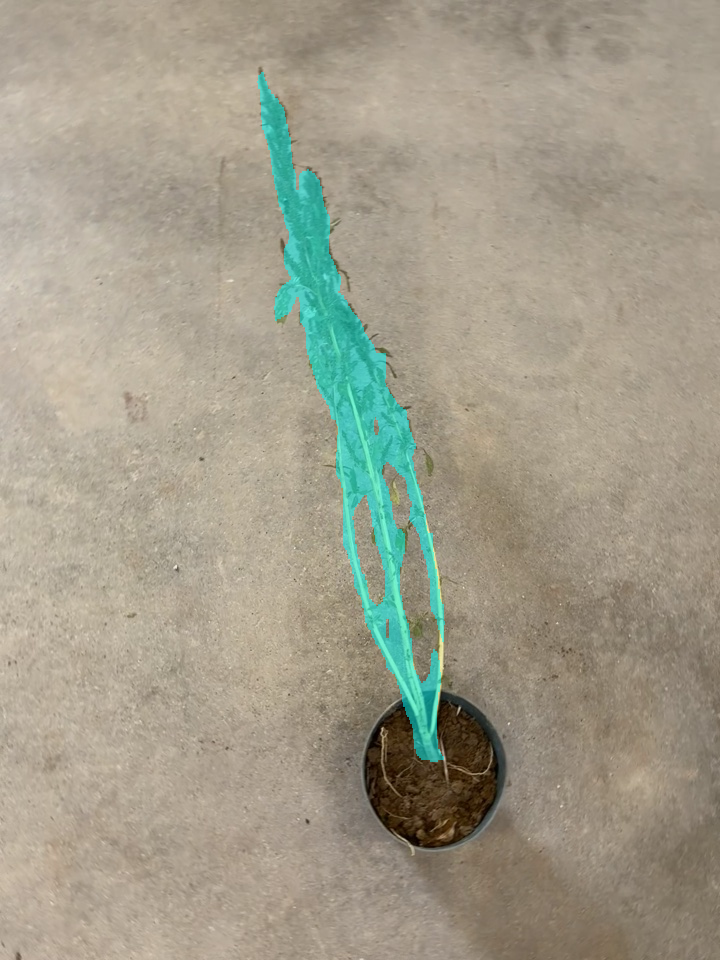}\hspace{\qualhspacing}
    \includegraphics[width=\qualfigsize\textwidth]{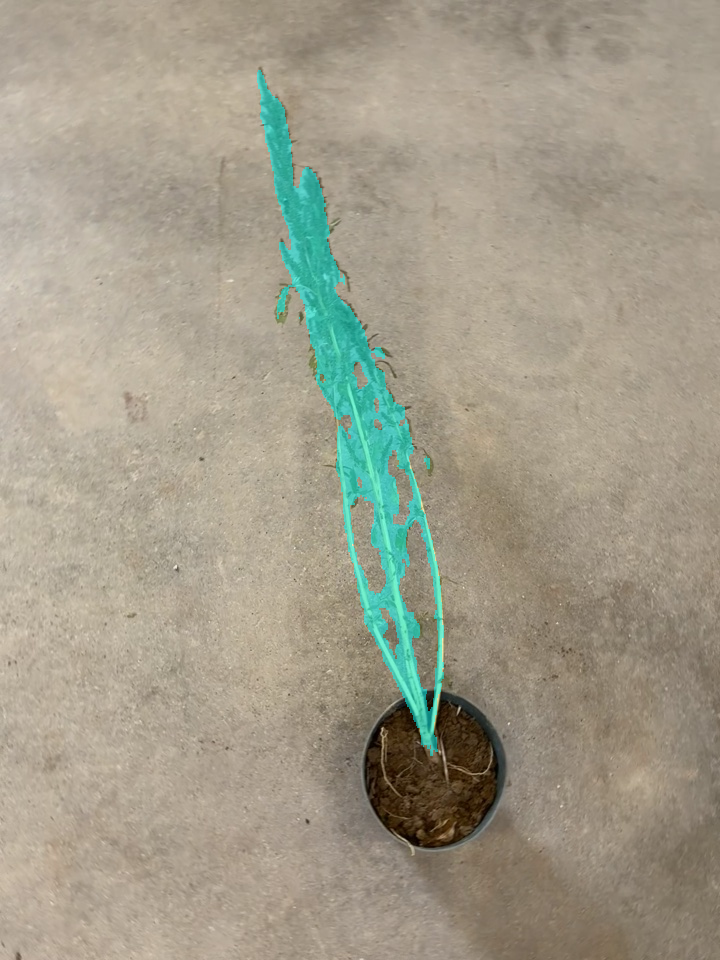}\hspace{\qualhspacing}
    \includegraphics[width=\qualfigsize\textwidth]{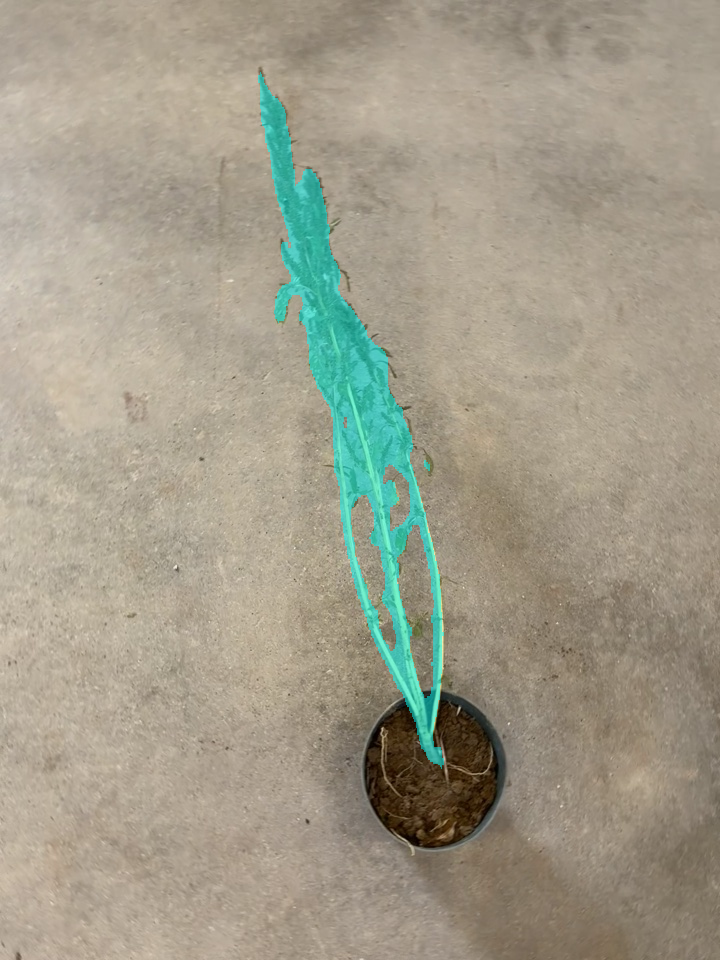}\hspace{\qualhspacing}
    \includegraphics[width=\qualfigsize\textwidth]{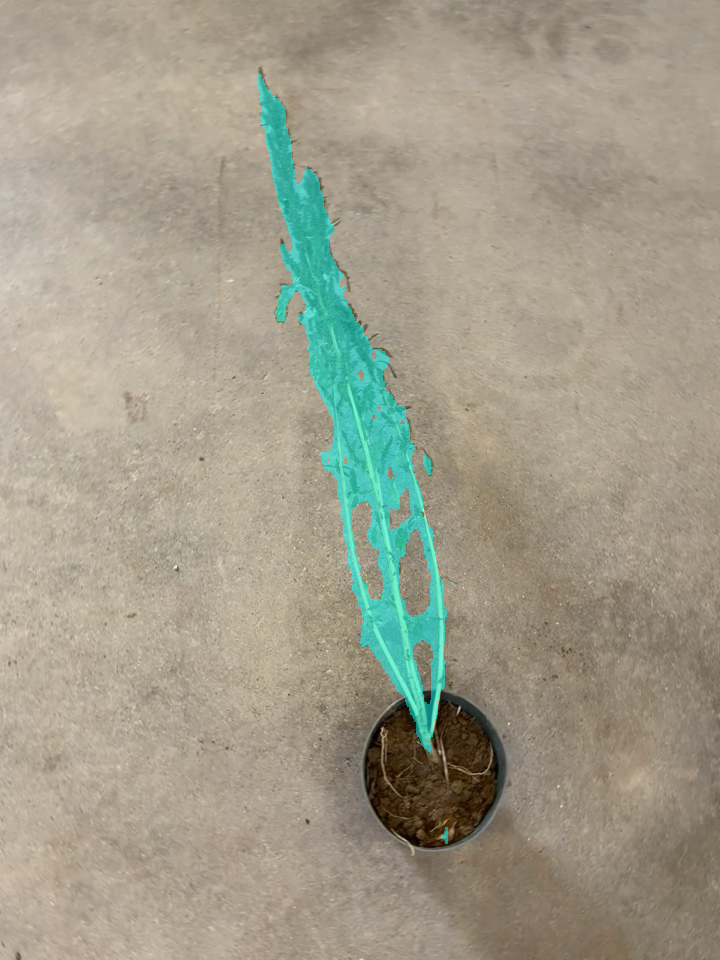}

    \caption{Qualitative comparison on male and female waterhemp. From left to right: input, ground truth, RepViT, FastViT, MobileOne, MobileNetV2, MobileNetV3, SegFormer-B0, and WeedRepFormer (ours). Best viewed on screen.}
    \label{fig:qualitative-comparison}
    \vspace{-15pt}
\end{figure*}

\begin{table}[b]
\centering
\scriptsize
\setlength{\tabcolsep}{2.5pt}
\begin{tabular}{@{}l|cc|cc|cc|cc@{}}
\toprule
\multirow{4}{*}{\textbf{Method}} & \multicolumn{4}{c|}{\textbf{Segmentation}} & \multicolumn{4}{c}{\textbf{Classification}} \\
\cmidrule(lr){2-5} \cmidrule(lr){6-9}
& \multicolumn{2}{c|}{\textbf{IoU (\%)$\uparrow$}} & \multicolumn{2}{c|}{\textbf{F-score (\%)$\uparrow$}} & \multicolumn{2}{c|}{\textbf{Acc (\%)$\uparrow$}} & \multicolumn{2}{c}{\textbf{F1 (\%)$\uparrow$}} \\
\cmidrule(lr){2-3} \cmidrule(lr){4-5} \cmidrule(lr){6-7} \cmidrule(lr){8-9}
 & \textbf{BG} & \textbf{Weed} & \textbf{BG} & \textbf{Weed} & \textbf{Male} & \textbf{Female} & \textbf{Male} & \textbf{Female} \\
\midrule
\multicolumn{9}{l}{\textit{Conv-based Models}} \\
\midrule
BiSeNet & 98.48 & 86.06 & 99.23 & 92.51 & 81.34 & 54.29 & 70.41 & 63.33 \\
BiSeNet V2 & 98.51 & 86.10 & 99.25 & 92.53 & 68.08 & 49.06 & 60.93 & 55.00 \\
DDRNet & 98.63 & 86.99 & 99.31 & 93.05 & 60.35 & 32.31 & 51.59 & 38.28 \\
Fast-SCNN & 97.59 & 78.95 & 98.78 & 88.24 & 55.69 & 63.14 & 56.89 & 61.93 \\
ICNet & 98.34 & 84.82 & 99.16 & 91.79 & 62.10 & 50.54 & 57.53 & 54.52 \\
MobileNetV2 & 98.34 & 84.66 & 99.16 & 91.69 & \textbf{97.38} & 46.51 & 76.21 & 62.47 \\
MobileNetV3 & 98.51 & 86.05 & 99.25 & 92.50 & 91.11 & 48.66 & 73.79 & 62.05 \\
MobileOne-S0 & 93.56 & 52.49 & 96.67 & 68.84 & 82.51 & 40.48 & 66.75 & 51.71 \\
RepViT-M0.9 & \textbf{98.93} & \textbf{89.69} & \textbf{99.46} & \textbf{94.56} & 83.24 & 56.30 & 72.14 & 65.57 \\
\midrule
\multicolumn{9}{l}{\textit{Transformer-based Models}} \\
\midrule
SegFormer-B0 & 98.88 & 89.31 & 99.44 & 94.36 & 90.09 & 60.59 & 77.35 & 71.41 \\
iFormer-T & 98.87 & 89.24 & 99.43 & 94.31 & 95.19 & 61.26 & 80.22 & 73.95 \\
FastViT-T8 & 98.73 & 87.93 & 99.36 & 93.58 & 71.72 & 40.08 & 60.55 & 48.26 \\
\midrule
\rowcolor{gray!20}
\textbf{WeedRepFormer} & 98.45 & 85.91 & 99.22 & 92.42 & 88.05 & \textbf{76.27} & 82.34 & \textbf{81.46} \\
\bottomrule
\end{tabular}
\vspace{-5pt}
\caption{Class-wise performance breakdown. WeedRepFormer achieves the best female classification accuracy (76.27\%) and F1-score (81.46\%), with the smallest male-female accuracy gap (11.78\%) compared to other methods.}
\label{tab:classwise}
\vspace{-10pt}
\end{table}

\noindent\textbf{Efficiency Analysis.} Table~\ref{tab:comparison} highlights the efficiency advantages of WeedRepFormer. With only 3.59M parameters and 3.80 GFLOPs, our model achieves 108.95 FPS while maintaining high multi-task accuracy (92.18\% mIoU, 81.91\% mAcc). Compared to the widely used SegFormer-B0 (7.89 GFLOPs, 115.27 FPS), WeedRepFormer reduces computational cost by 51.7\% and parameter count by 5.0\% with negligible impact on throughput. While Fast-SCNN is more lightweight (0.93 GFLOPs), it suffers significant accuracy degradation. Conversely, higher-capacity models like iFormer-T (24.27 GFLOPs) and RepViT-M0.9 (25.40 GFLOPs) incur $6.4\times$ and $6.7\times$ higher computational costs, respectively. Furthermore, our structural reparameterization strategy yields a $1.54\times$ inference speedup, collapsing from 4.63 GFLOPs (70.89 FPS) during training to 3.80 GFLOPs (108.95 FPS) at deployment. These results position WeedRepFormer as an optimal solution for resource-constrained agricultural robotics.

\noindent\textbf{Class-wise Performance Analysis.} Table~\ref{tab:classwise} provides a detailed breakdown of per-class performance. For segmentation, all methods achieve near-saturated background IoU ($>$97\%), with primary differentiation occurring on the weed class where RepViT-M0.9 leads at 89.69\% IoU. For classification, most methods exhibit severe male-female accuracy disparities. MobileNetV2 achieves the highest male accuracy yet drops to only 46.51\% on females, resulting in a gap of 50.87\%. Similarly, iFormer-T and DDRNet show gaps of 33.93\% and 28.04\%, respectively. In contrast, WeedRepFormer achieves the best female accuracy of 76.27\% and F1-score of 81.46\% with the smallest gap of only 11.78\%. This balanced performance is agriculturally significant: female waterhemp plants produce seeds and drive herbicide resistance spread, making their reliable detection essential for effective weed management.

\begin{table}[t]
\centering
\scriptsize
\setlength{\tabcolsep}{3pt}
\begin{tabular}{@{}l|cc|ccc@{}}
\toprule
\textbf{Configuration} & \textbf{\makecell{mIoU\\(\%)$\uparrow$}} & \textbf{\makecell{mAcc\\(\%)$\uparrow$}} & \textbf{\makecell{FPS$\uparrow$}} & \textbf{\makecell{GFLOPs$\downarrow$}} & \textbf{\makecell{Params\\(M)$\downarrow$}} \\
\midrule
\multicolumn{6}{l}{\textit{Reparameterizable Backbone Components}} \\
\midrule
RepMixFFN & 92.76 & 62.85 & 112.02 & 3.797 & 3.582 \\
RepPatchEmbed & 91.93 & 61.24 & 114.58 & 3.797 & 3.582 \\
RepCPE & 93.74 & 68.09 & 105.02 & 3.801 & 3.590 \\
RepCPE + RepMixFFN & 92.72 & 52.72 & 111.60 & 3.801 & 3.590 \\
RepCPE + RepPatchEmbed & 91.66 & 66.27 & 109.67 & 3.801 & 3.590 \\
RepMixFFN + RepPatchEmbed & 91.89 & 73.95 & 117.35 & 3.797 & 3.582 \\
\rowcolor{gray!20}
All Components & 92.15 & 77.09 & 109.79 & 3.801 & 3.590 \\
\midrule
\multicolumn{6}{l}{\textit{Classification Head Design}} \\
\midrule
$K$=1, Standard MLP & 92.15 & 77.09 & 109.79 & 3.801 & 3.590 \\
$K$=1, RepClsHead & 92.32 & 79.19 & 109.99 & 3.801 & 3.592 \\
$K$=2, Standard MLP & 92.25 & 77.72 & 109.26 & 3.801 & 3.590 \\
\rowcolor{gray!20}
$K$=2, RepClsHead & 92.18 & \textbf{81.91} & \textbf{108.95} & 3.801 & 3.592 \\
\midrule
\multicolumn{6}{l}{\textit{Parallel Branch Count ($K$)}} \\
\midrule
$K$=1, RepClsHead & 92.32 & 79.19 & 109.99 & 3.801 & 3.592 \\
\rowcolor{gray!20}
$K$=2, RepClsHead & 92.18 & 81.91 & 108.95 & 3.801 & 3.592 \\
$K$=3, RepClsHead & 92.25 & 75.56 & 108.31 & 3.801 & 3.592 \\
$K$=4, RepClsHead & 92.17 & 78.84 & 108.26 & 3.801 & 3.592 \\
\midrule
\multicolumn{6}{l}{\textit{RepPatchEmbed Factorization (V2: Regular Conv vs V3: Depthwise-Pointwise)}} \\
\midrule
V2, $K$=1, Standard MLP & 92.15 & 77.09 & 109.79 & 3.801 & 3.590 \\
V2, $K$=1, RepClsHead & 92.32 & 79.19 & 109.99 & 3.801 & 3.592 \\
\rowcolor{gray!20}
V2, $K$=2, RepClsHead & 92.18 & 81.91 & 108.95 & 3.801 & 3.592 \\
V3, $K$=1, Standard MLP & 90.14 & 59.01 & 110.15 & \textbf{3.495} & \textbf{3.162} \\
V3, $K$=1, RepClsHead & 90.30 & 59.01 & 109.45 & 3.496 & 3.165 \\
V3, $K$=2, RepClsHead & 88.58 & 56.22 & 108.42 & 3.496 & 3.165 \\
\midrule
\multicolumn{6}{l}{\textit{CPE Placement Across Stages (Stage1, Stage2, Stage3, Stage4)}} \\
\midrule
(T, T, T, T) & 91.98 & 69.90 & 106.27 & 3.816 & 3.594 \\
(T, T, F, F) & 92.03 & 65.78 & 110.08 & 3.811 & 3.586 \\
\rowcolor{gray!20}
(F, F, T, T) & 92.18 & 81.91 & 108.95 & 3.801 & 3.592 \\
(T, F, T, F) & 92.06 & 68.09 & 105.58 & 3.810 & 3.588 \\
(F, T, F, T) & 92.35 & 73.46 & 109.88 & 3.803 & 3.591 \\
\midrule
\multicolumn{6}{l}{\textit{RepPatchEmbed Kernel Size Configuration}} \\
\midrule
\rowcolor{gray!20}
(7, 3, 3, 3) & 92.18 & 81.91 & 108.95 & 3.801 & 3.592 \\
(7, 7, 7, 7) & 92.08 & 81.08 & 109.01 & 4.976 & 5.722 \\
\midrule
\multicolumn{6}{l}{\textit{Squeeze-and-Excitation in Classification Head}} \\
\midrule
\rowcolor{gray!20}
Without SE & 92.18 & 81.91 & 108.95 & 3.801 & 3.592 \\
With SE & 92.12 & 76.19 & 109.03 & 3.801 & 3.601 \\
\bottomrule
\end{tabular}
\vspace{-5pt}
\caption{Comprehensive ablation study of architectural components. Gray rows indicate selected configurations. Results demonstrate systematic optimization from individual reparameterizable components through branch count selection, factorization strategy, CPE placement, patch size configuration, and attention mechanisms.}
\label{tab:comprehensive_ablation}
\vspace{-10pt}
\end{table}

\noindent\textbf{Qualitative Results.} Figure~\ref{fig:qualitative-comparison} illustrates segmentation results on representative male and female Waterhemp samples. MobileOne produces degraded segmentation with coarse boundaries and missed regions. Other methods achieve visually similar segmentation quality. WeedRepFormer produces comparable visual quality to higher-capacity models like RepViT-M0.9 and SegFormer-B0, validating that structural reparameterization enables effective multi-task learning without compromising segmentation quality.

\subsection{Ablation Study}

We conduct a systematic ablation to validate architectural decisions. All experiments utilize the MixVisionTransformer backbone and RepLR-ASPP decoder unless otherwise specified. Results are summarized in Table~\ref{tab:comprehensive_ablation}.

\noindent\textbf{Reparameterizable Backbone Components.} We first evaluate the impact of including RepMixFFN, RepPatchEmbed, and RepCPE in the backbone. As shown in Table~\ref{tab:comprehensive_ablation}, using RepCPE alone achieves the highest segmentation performance (93.74\% mIoU) but yields suboptimal classification accuracy (68.09\% mAcc). While individual and pairwise combinations show varying trade-offs, combining all three components achieves the best multi-task balance, yielding 92.15\% mIoU and 77.09\% mAcc. This configuration significantly improves classification (+9.0\% mAcc over RepCPE alone) with only a minor trade-off in segmentation, validating that multi-component reparameterization is essential for capturing both the spatial context and semantic features required for simultaneous segmentation and classification.

\noindent\textbf{Classification Head Design.} We challenge the standard MLP head design. Replacing the Standard MLP with our proposed RepClsHead yields immediate gains. Even with a single branch ($K=1$), accuracy improves by +2.1\%. With $K=2$ branches, RepClsHead achieves 81.91\% mAcc versus 77.72\% for the simple baseline. This validates that the classification head benefits significantly from the increased capacity of multi-branch training, which is subsequently compressed for inference.

\noindent\textbf{Branch Count ($K$) \& Efficiency.} We investigate the training-time width of the reparameterizable blocks ($K \in \{1,2,3,4\}$). We observe a performance peak at $K=2$. Increasing branches further to $K=3$ or $K=4$ leads to diminishing returns and overfitting, degrading accuracy to 75-78\%. The $K=2$ configuration offers the optimal sweet spot: it incurs computational cost only during training, collapsing to the exact same inference cost (3.80 GFLOPs) as the $K=1$ baseline, effectively providing a +2.7\% accuracy boost via structural reparameterization.

\noindent\textbf{Patch Embedding Factorization.} We evaluate depthwise-pointwise factorization (V3) against regular convolution (V2) within RepPatchEmbed. While V3 offers a theoretical reduction in GFLOPs (3.80 $\rightarrow$ 3.50), it causes a catastrophic drop in classification accuracy ($>20\%$ decline). This finding suggests that dense feature interaction, provided by regular convolutions, is critical for capturing the subtle visual cues required for weed gender classification.

\noindent\textbf{Conditional Positional Encoding Placement.} We analyze the optimal insertion points for Conditional Positional Encodings (CPE) across the four backbone stages. We find that applying RepCPE in early stages is detrimental to classification performance. The best balance is achieved by restricting RepCPE to the deeper layers (Stages 3 \& 4). This aligns with the intuition that early stages focus on local, translation-invariant texture features, whereas deeper stages capture high-level semantic information where conditional spatial context is most beneficial.

\noindent\textbf{RepPatchEmbed Kernel Size Configuration.} We investigate the impact of kernel size scaling within the embedding modules. We compare a uniform large-kernel strategy [7, 7, 7, 7] against a progressive reduction strategy [7, 3, 3, 3]. The uniform configuration incurs a significant parameter penalty (5.72M) without improving performance. In contrast, the progressive strategy achieves higher accuracy with only 3.59M parameters. This confirms that while a large initial receptive field is crucial for early tokenization, subsequent stages benefit from efficient, smaller kernels to refine features without unnecessary computational overhead.

\noindent\textbf{Squeeze-and-Excitation Attention.} Finally, we investigated the integration of Squeeze-and-Excitation (SE) attention~\cite{hu2018squeeze} within the RepClsHead to potentially enhance feature selection. Contrary to expectations, adding SE resulted in a 5.72\% degradation in classification accuracy. Given this significant performance penalty, we exclude attention mechanisms from the final architecture.

%% file: sec/06_conclusion.tex
\section{Conclusion}
This paper introduces WeedRepFormer, which systematically integrates structural reparameterization throughout a Vision Transformer architecture for simultaneous waterhemp segmentation and gender classification. Unlike prior agricultural work that primarily uses CNN-based reparameterization, we apply it to hierarchical Vision Transformers across backbone, decoder, and task-specific heads for joint dense prediction and gender classification. WeedRepFormer outperforms all compared methods in classification accuracy while maintaining competitive segmentation with significantly reduced computational cost. Notably, our model achieves the best female classification performance with the smallest male-female accuracy gap, addressing a critical performance disparity. From an agricultural perspective, reliable female plant detection enables targeted weed management since female waterhemp are the primary seed producers responsible for population spread and herbicide resistance.